\documentclass[12pt, a4paper]{article}

\usepackage[margin=1in]{geometry}  
\usepackage{times}                 

\usepackage{amsmath,amssymb,amsfonts}
\usepackage{graphicx}
\usepackage{setspace} 
\usepackage{subcaption}
\usepackage{indentfirst}
\usepackage{xurl}

\usepackage[style=apa, backend=biber]{biblatex}
\usepackage[colorlinks=true, allcolors=blue]{hyperref}

\AtBeginBibliography{\normalsize} 
\AtBeginBibliography{\singlespacing}

\usepackage{titlesec}
\titleformat{\section}{\normalfont\normalsize\bfseries}{\thesection.}{1em}{}
\titleformat{\subsection}{\normalfont\normalsize\itshape}{\thesubsection.}{1em}{}
\titleformat{\subsubsection}{\normalfont\normalsize\itshape}{\thesubsubsection.}{1em}{}

\usepackage{caption}
\begin{document}

\noindent
{\large \bfseries MatchPlant: An Open-Source Pipeline for UAV-Based Single-Plant \\Detection and Data Extraction}\\[1.5em]
{\small Worasit Sangjan\textsuperscript{a}, Piyush Pandey\textsuperscript{a}, Norman B. Best\textsuperscript{a}, Jacob D. Washburn\textsuperscript{a,*}}\\[0.5em]
{\small \textsuperscript{a}USDA-ARS, Plant Genetics Research Unit, Columbia, MO, United States}\\[1em]
{\footnotesize \textsuperscript{*}Corresponding author.\\
Email address: \href{mailto:jacob.washburn@usda.gov}{jacob.washburn@usda.gov}}
\vspace{2em}

\noindent 
\textbf{Abstract:} 
\begin{spacing}{1.5}
Accurate identification of individual plants from unmanned aerial vehicle (UAV) images is essential for advancing high-throughput phenotyping and supporting data-driven decision-making in plant breeding. This study presents MatchPlant, a modular, graphical user interface-supported, open-source Python pipeline for UAV-based single-plant detection and geospatial trait extraction. MatchPlant enables end-to-end workflows by integrating UAV image processing, user-guided annotation, Convolutional Neural Network model training for object detection, forward projection of bounding boxes onto an orthomosaic, and shapefile generation for spatial phenotypic analysis. In an early-season maize case study, MatchPlant achieved reliable detection performance (validation AP: 89.6\%, test AP: 85.9\%) and effectively projected bounding boxes, covering 89.8\% of manually annotated boxes with 87.5\% of projections achieving an Intersection over Union (IoU) greater than 0.5. Trait values extracted from predicted bounding instances showed high agreement with manual annotations ($r$ = 0.87--0.97, IoU $\geq$ 0.4). Detection outputs were reused across time points to extract plant height and Normalized Difference Vegetation Index with minimal additional annotation, facilitating efficient temporal phenotyping. By combining modular design, reproducibility, and geospatial precision, MatchPlant offers a scalable framework for UAV-based plant-level analysis with broad applicability in agricultural and environmental monitoring.
\vspace{2em}

\noindent
\textbf{Keywords:} UAV phenotyping, graphical user interface (GUI), Faster R-CNN, object detection, plant breeding, geospatial analysis, orthomosaic projection
\vspace{2em} 
\pagebreak 
\end{spacing}

\begin{spacing}{2}
\section{Introduction}
High-throughput imaging technologies are transforming modern agriculture by integrating advanced imaging techniques, diverse sensors, and automated data processing. These tools—including satellites, unmanned aerial vehicles (UAVs), robots, Internet of Things (IoT) systems, and various sensors integrated with artificial intelligence—enable the precise and efficient measurement of plant traits, livestock behavior, and environmental conditions. Applications span agricultural research, precision crop management, and pre- and post-harvest monitoring, all contributing to enhanced productivity and sustainability \autocite{kganyago2024optical, rui2024high, sangjan2023evaluation}.

Object detection, a key application in computer vision, involves identifying and localizing objects within images or videos using deep learning (DL) techniques. Methods such as Faster Region-based Convolutional Neural Network (Faster R-CNN); \cite{ren2015faster}), Mask R-CNN \autocite{he2017mask}, and You Only Look Once (YOLO; \cite{redmon2016you}) have demonstrated remarkable effectiveness in agricultural research and practice. These methods facilitate various applications, including the detection and classification of individual plants, fruits, and vegetables \autocite{gimenez2024tree, zhuang2024maize}, weed identification and management \autocite{cui2024weed, rai2024weedvision}, pest and disease detection \autocite{zhang2024using, zhu2024research}, and livestock tracking and monitoring \autocite{myint2024development, saeidifar2024zero}. By integrating these approaches into agricultural workflows, researchers and practitioners can optimize resource use, enhance yield potential, and promote sustainable practices.

Despite the promise of object detection models, several challenges hinder their broader adoption in agriculture. A key challenge lies in bridging the disciplinary divide between agricultural science and the technical expertise required for effective model deployment. Object detection pipelines demand multidisciplinary proficiency in data preprocessing, model training, hyperparameter optimization, and workflow integration within agricultural systems \autocite{huang2023survey, ariza2024object, rohan2024application}. Although several object detection tools exist, many require computer programming expertise or cloud-based access, which may limit adoption by researchers without a strong technical background. Another critical challenge is the availability of high-quality annotated datasets, which are essential for training robust models but require substantial time and resources to develop \autocite{checola2024novel, lu2024mar}.

Modular preprocessing workflows and open-source solutions are needed to address these challenges, equipping researchers with accessible and practical tools. UAV imagery offers high spatial resolution, flexible deployment, and rapid data acquisition, making it especially suitable for field-based phenotyping compared to other remote sensing platforms, such as satellite and ground-based imagery \autocite{rejeb2022drones, sangjan2024effect, gano2024drone}. However, a key limitation arises when using an orthomosaic—a composite of orthorectified UAV images—for training DL models. Orthorectification can introduce artifacts, distort object geometry, and stretch or shrink plant outlines, ultimately reducing detection precision \autocite{zheng2023object, lu2024weed}. 

This study presents MatchPlant, an open-source pipeline designed to streamline UAV-based object detection in agricultural settings. The pipeline integrates UAV image processing, data preparation, object detection model training and deployment, projection of detected plants, and geospatial trait extraction within a modular framework. Users can select individual components for specific tasks or integrate them into a complete workflow. The system leverages OpenDroneMap (ODM), an open-source platform that generates undistorted images, orthomosaics, and other outputs while providing features comparable to commercial software. ODM’s capabilities enable seamless access to both processed and raw imagery, supporting flexible and precise object detection workflows \autocite{opendronemap2020, gbagir2023opendronemap, valluvan2023canopy}. A key innovation in MatchPlant is training models directly on undistorted images rather than an orthomosaic to preserve spatial precision. Detected objects are then projected onto the orthomosaic, mitigating the geometric distortions introduced by orthorectification and improving detection accuracy. 

This research aims to develop an accessible pipeline that enables users to detect and delineate objects in large-scale agricultural contexts with spatial precision. While this study focuses on plant phenotyping as a case study, the proposed pipeline is potentially adaptable to various applications, including livestock monitoring and environmental assessments. Through this open-source system, which features intuitive user interfaces throughout the workflow, MatchPlant aims to bridge the gap between object detection technologies and their deployment as field-ready tools for agricultural research. A maize case study was conducted using UAV imagery from a genetic research field, validating the pipeline’s modular components for consistent detection, effective spatial projection, and trait-level data extraction.

\section{Pipeline Development}

\begin{figure}[htbp]
  \centering
  \includegraphics[width=\textwidth, height=\textheight, keepaspectratio]{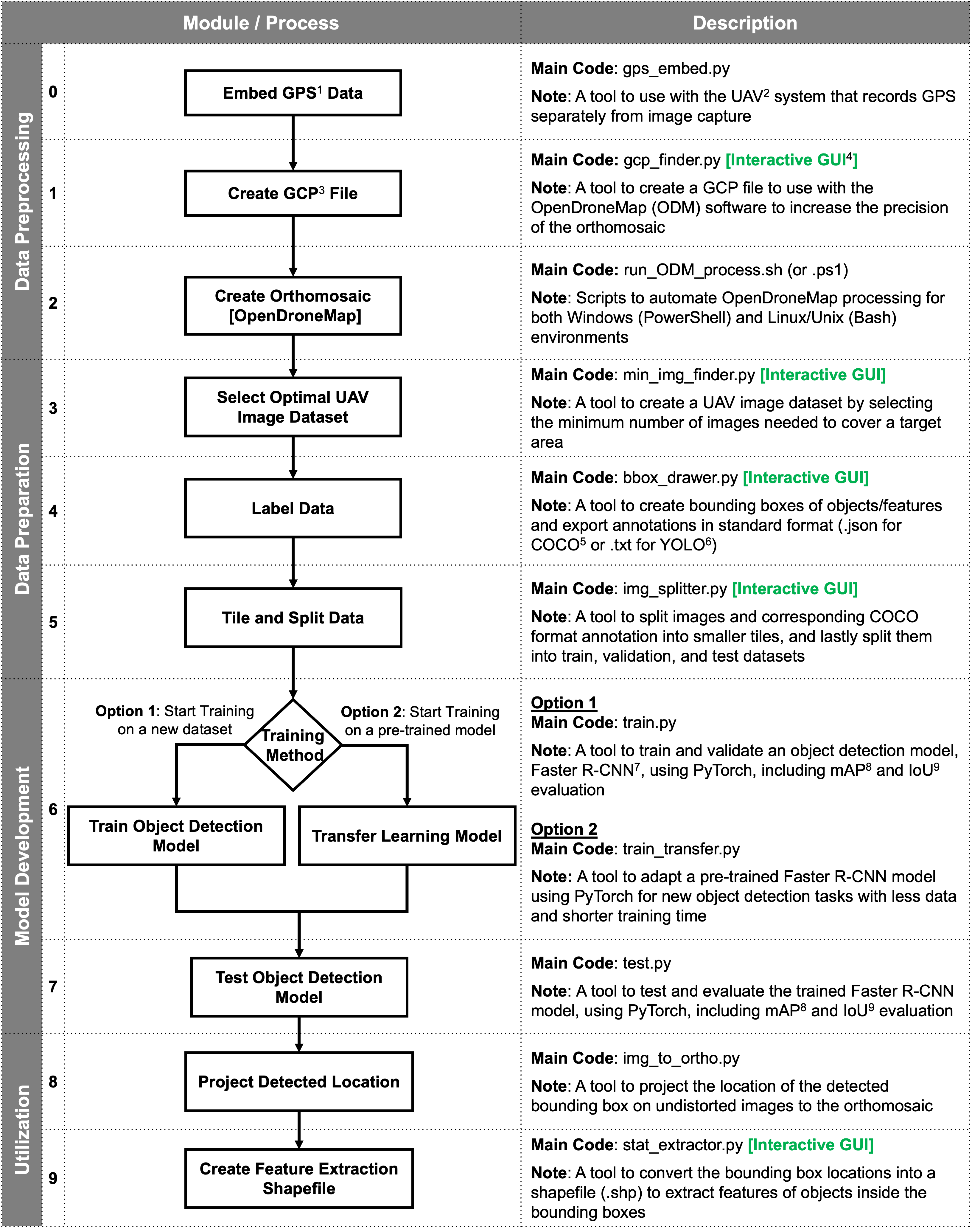}
  \caption{Modular overview of MatchPlant. All modules are publicly available at GitHub: \url{https://github.com/JacobWashburn-USDA/MatchPlant}; $^{1}$GPS: Global Positioning System, $^{2}$UAV: Unmanned Aerial Vehicle, $^{3}$GCP: Ground Control Point, $^{4}$GUI: Graphical User Interface, $^{5}$COCO: Common Objects in Context, $^{6}$YOLO: You Only Look Once, $^{7}$Faster R-CNN: Faster Region-based Convolutional Neural Network.}
  \label{fig:pipeline}
\end{figure}

The MatchPlant pipeline is a modular and adaptable system designed to detect and track individual objects in UAV imagery. It comprises four primary components: UAV image data preprocessing, dataset preparation for object detection, model development, and post-detection utilization (\autoref{fig:pipeline}). Most modules are implemented in Python 3 (\url{https://www.python.org/}, accessed on February 14, 2025), while the command-line module (Module 2) uses platform-specific execution scripts (*.sh for Linux and macOS and *.ps1 for Windows) to ensure broad compatibility. MatchPlant supports both Windows and macOS platforms and incorporates a combination of graphical user interface (GUI) modules into the workflows. GUI-based modules (1, 3, 4, 5, and 9) are customized for their respective operating systems to ensure cross-platform usability.

Model development is facilitated through standardized Python scripts for training (Module 6.1), transfer learning (Module 6.2), and testing (Module 7). Each script is configured via YAML (YAML Ain’t Markup Language), allowing users to customize hyperparameters, model architecture, and dataset paths without modifying the core code. These scripts feature built-in hardware detection to automatically configure the appropriate computation backend, leveraging Compute Unified Device Architecture (CUDA) for graphics processing unit (GPU) acceleration on Windows and Linux systems and Metal Performance Shaders (MPS) on macOS. 

The MatchPlant codebase and documentation are publicly available on GitHub at \url{https://github.com/JacobWashburn-USDA/MatchPlant} (accessed on February 14, 2025). Each module includes a dedicated README file describing system requirements, Python dependencies, input/output formats, and detailed execution instructions to support technical and non-technical users. The training dataset and pre-trained model used in the maize case study presented in Section 3 are also publicly available via Zenodo \autocite{matchplant2025dataset} at \url{https://doi.org/10.5281/zenodo.14856123} (accessed on February 14, 2025). 

\subsection{Data Preprocessing}
The data preprocessing stage prepares UAV imagery for object detection by ensuring accurate georeferencing and standardized formatting. This includes embedding a global positioning system (GPS), creating a ground control point (GCP) file, and UAV image data processing using ODM. These steps establish a geospatially reliable foundation for downstream model training and trait extraction.

\subsubsection{Module 0: GPS Embedding}
Some UAV platforms record global navigation satellite system (GNSS) data separately rather than embedding GPS metadata directly into image files. Module 0, “0\_gps\_embeder,” synchronizes image timestamps with GNSS records using a nearest-neighbor matching algorithm. Matched coordinates are transformed from Universal Transverse Mercator (UTM) to World Geodetic System 1984 (WGS84) using PyProj (\url{https://pypi.org/project/pyproj/}, accessed on February 14, 2025) and written into the image’s Exchangeable Image File Format (EXIF) metadata. 

\subsubsection{Module 1: GCP File Creation}
\begin{figure}[htbp]
  \centering
  \includegraphics[width=\textwidth, height=\textheight, keepaspectratio]{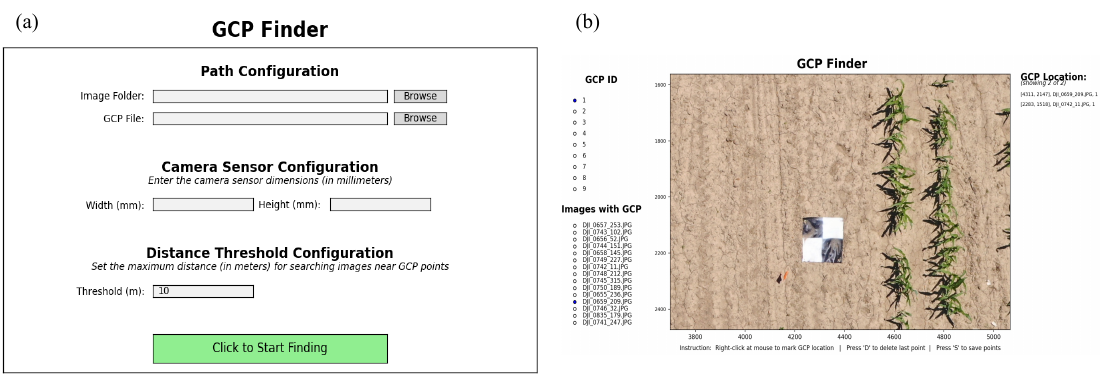}
  \caption{GUI interface of the “1\_gcp\_finder” for marking GCPs. (a) Configuration interface for selecting an image folder, GCP files, and setting sensor parameters. (b) The main GUI is showing GCP candidates for refinement and confirmation.}
  \label{fig:mod1}
\end{figure}

GCPs are critical for accurate orthorectification. Module 1, “1\_gcp\_finder,” automates the identification of UAV images that contain visible GCPs and generates a GCP list file (gcp\_list.txt) for input into the ODM workflow. The module filters images based on spatial proximity to known GCP locations. An interactive Matplotlib-based (\url{https://matplotlib.org/}, accessed on February 14, 2025) GUI is developed for manual marking and validating GCP positions (\autoref{fig:mod1}). Image display and processing are handled with OpenCV (\url{https://opencv.org/}, accessed on February 14, 2025).

After generating the GCP list, the validated coordinates are converted from WGS84 to UTM for compatibility with geospatial tools. The accurate GCP localization improves alignment, enhancing orthomosaic quality and geospatial accuracy.

\subsubsection{Module 2: UAV Image Preprocessing}
Module 2, “2\_odm\_runner,” automates UAV RGB (Red, Green, Blue) image processing using ODM via Docker (\url{https://www.docker.com/}, accessed on February 14, 2025). If a GCP file from Module 1 is available, it is incorporated into the processing workflow. ODM outputs include orthomosaic, orthorectified images, and undistorted images, which are input for subsequent object detection steps. The orthomosaic provides a stitched, georeferenced top-down view of the field, while orthorectified images correct camera and terrain-induced distortions. Undistorted images retain the camera’s native perspective with lens correction, preserving object integrity for applications that require raw but corrected imagery. 

ODM installation and setup instructions are available at \url{https://github.com/OpenDroneMap/ODM} (accessed on February 14, 2025).

\subsection{Data Preparation}
The data preparation stage ensures that UAV imagery is formatted correctly, annotated, and structured for object detection model training. This step involves selecting relevant images, drawing bounding boxes, and splitting datasets to form a standardized input for DL models.

\subsubsection{Module 3: Image Selection}
\begin{figure}[htbp]
  \centering
  \includegraphics[width=\textwidth, height=\textheight, keepaspectratio]{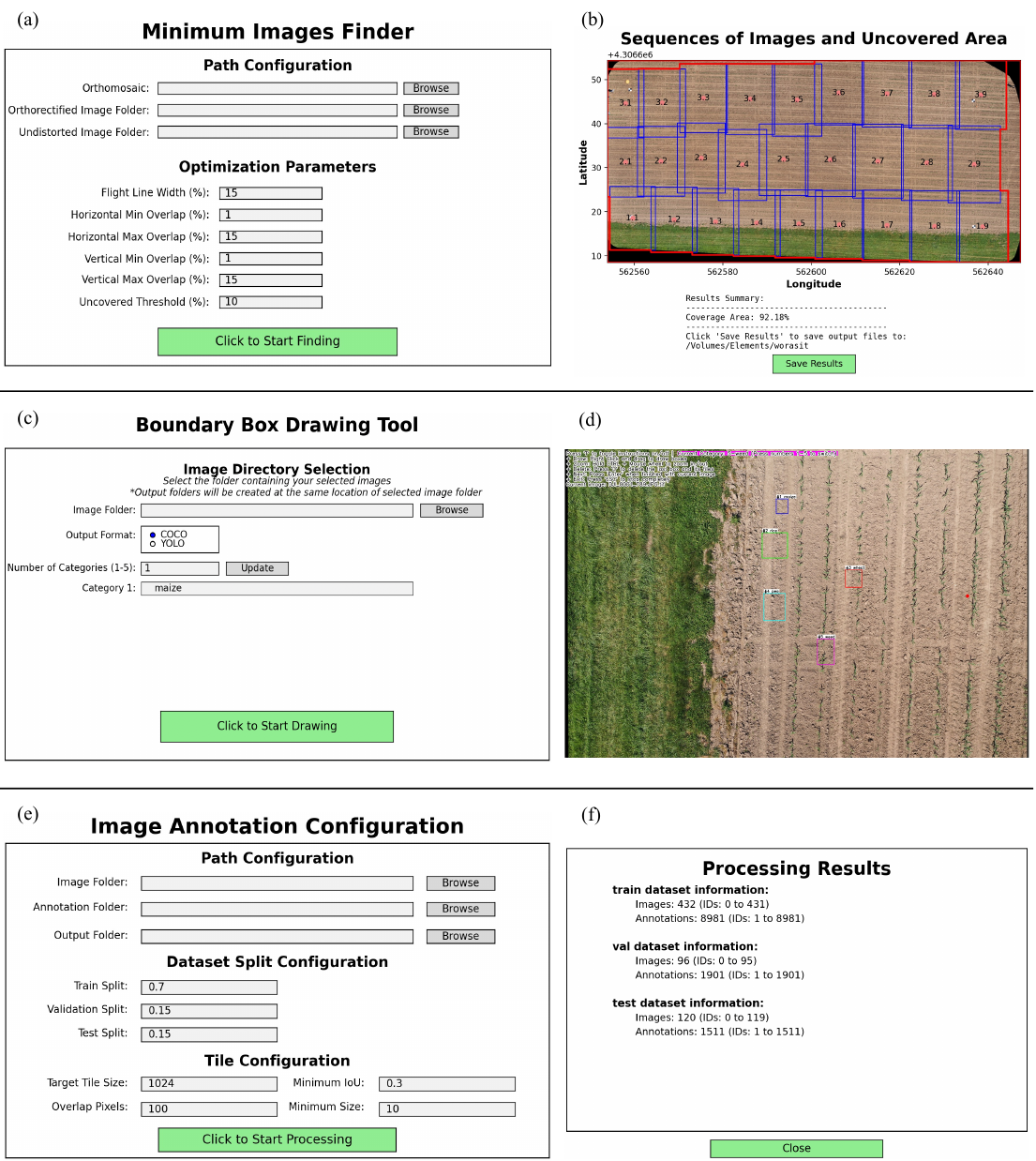}
  \caption{GUIs for the data preparation stage include image selection, bounding box annotation, and dataset splitting. (a) The GUI for the “3\_min\_img\_finder” module enables users to select the minimum set of images based on spatial coverage criteria. (b) Visualization illustrates the chosen image sequences and uncovered regions for validation. (c) The GUI for the “4\_bbox\_drawer” module facilitates manual bounding box annotation, supports labeling across five categories, and exports in COCO or YOLO formats. (d) Real-time visualization displays bounding box annotations overlaid on an image. (e) The GUI for the “5\_img\_splitter” module allows users to set the desired image size and divides the annotated dataset into training, validation, and test sets. (f) Processing results show the dataset is organized for the object detection model workflow.}
  \label{fig:mod345}
\end{figure}

The “3\_min\_img\_finder” module optimizes UAV datasets for object detection by selecting the minimum number of images necessary while maintaining full coverage of the area of interest. High image overlap is common in UAV-based plant phenotyping, often resulting in repeated capture of the same plants, redundant labels, and ineffective validation. To address this, the tool uses orthorectified UAV images and associated metadata to identify the most relevant undistorted images, reducing redundancy while ensuring complete spatial completeness. The selection algorithm considers flight line width, horizontal and vertical overlap thresholds, and uncovered area percentages. An interactive GUI built with Matplotlib and OpenCV allows users to configure selection parameters, visualize coverage, and validate the selected image set (\hyperref[fig:mod345]{Figures 3a and 3b}).

\subsubsection{Module 4: Bounding Box Annotation}
The “4\_bbox\_drawer” module provides an interactive GUI for manually annotating objects with bounding boxes, supporting up to five distinct object categories (\hyperref[fig:mod345]{Figures 3c and 3d}). It allows users to generate structured datasets for object detection and supports output formats in the common objects in context (COCO, *.json) and YOLO (*.txt) formats, ensuring compatibility with major DL frameworks. Developed using Matplotlib and OpenCV, the GUI enables users to zoom, pan, and modify bounding boxes. Real-time visualization allows users to validate annotations before saving, improving annotation accuracy. Final annotation files contain bounding box coordinates and associated metadata. 

\subsubsection{Module 5: Image Tiling and Splitting for Model Training}
This module segments annotated UAV images into smaller tiles and splits them into training, validation, and test sets to provide efficient processing and data handling. The module “5\_img\_splitter” includes a GUI for setting tile size, overlap control (default: 100 pixels), and dataset split ratios (\hyperref[fig:mod345]{Figures 3e and 3f}). A minimum Intersection over Union (IoU) threshold (default: 0.3) ensures only well-preserved objects are retained within each tile, minimizing object fragmentation at tile boundaries. Bounding boxes that fall below the overlap or size thresholds are excluded, ensuring the dataset includes only meaningful detections. This dynamic adjustment mechanism helps maintain annotation quality near tile edges. For COCO-format annotations, bounding box coordinates are recalculated to align with the newly created image tiles. The final dataset is structured for direct integration into an object detection model workflow, with dataset distribution governed by user-defined settings.

\subsection{Model Development}
This stage focuses on training and evaluating the object detection model. The framework offers two training approaches: training a model from the ground up or fine-tuning a pre-trained model using transfer learning. Training, transfer learning, and testing modules support multi-GPU acceleration using CUDA and MPS, enabling scalable object detection in UAV imagery.

\subsubsection{Module 6.1: Model Training}
Faster R-CNN is implemented as the core detection algorithm using PyTorch (\url{https://pytorch.org/}, accessed on February 14, 2025), tailored for UAV-based maize detection in early growth stages. The model is initialized with COCO pre-trained weights \autocite{lin2014microsoft} and uses a ResNet-50 backbone \autocite{he2016deep} with an FPN (Feature Pyramid Network; \cite{lin2017feature}) for multi-scale feature extraction. As a two-stage detector, it generates region proposals before classifying them, improving small object detection performance in UAV images \autocite{liu2021survey, koyun2022focus}. 

The training configuration is defined in “train\_config.yaml,” where users can adjust the key region proposal network (RPN), region of interest (ROI), and anchor settings. The training process uses a stochastic gradient descent (SGD) optimization with custom learning rates for the backbone, RPN, and ROI heads combined with a learning rate scheduler. Augmentation techniques such as horizontal flipping, brightness, contrast, and saturation adjustments are implemented to improve model generalization. During modeling, checkpointing, and validation, tracking is provided for performance feedback on classification loss, bounding box regression loss, average precision (AP) at multiple IoUs, and mean average precision (mAP)

\subsubsection{Module 6.2: Transfer Learning Model}
Module 6.2 enables fine-tuning of a pre-trained Faster R-CNN model using datasets formatted in COCO style. Users can configure the “transfer\_config.yaml” to adjust key parameters such as the number of classes, anchor sizes, and detection settings. Selective layer freezing helps retain generic features while optimizing target-specific layers. Different learning rates are applied to the backbone, RPN, and ROI heads for controlled adaptation. The pre-trained and target models must share the same architecture to ensure compatibility, and datasets must be formatted in COCO style for seamless integration.

\subsubsection{Module 7: Model Testing}
The testing pipeline is designed to maintain consistency with the training framework by supporting datasets structured in the COCO annotation format. Configurable parameters in “test\_config.yaml” allow users to define evaluation settings, including the number of test iterations, confidence thresholds, and IoU thresholds. The evaluation process measures precision, recall, F1 score, and mAP across small, medium, and large object size categories.

Results are logged in structured formats, enabling detailed performance analysis. The module facilitates saving both individual and aggregated outputs, including raw predictions and graphical visualizations of detected objects. 

\subsection{Utilization}
The final stage of the pipeline incorporates detected plant locations into geospatial analyses. Bounding boxes from detected plants are projected onto the corresponding orthomosaic to align plant positions with geospatial reference data. Additionally, spatial data conversion transforms projected outputs into shapefiles, enabling integration with geographic information system (GIS)-based analysis. These processes improve the usability of UAV-derived plant detection for agronomic and breeding research, supporting applications in crop monitoring, trait analysis, and decision-making.

\subsubsection{Module 8: Detected Plant Projection}
Module 8 applies forward projection to transfer bounding box coordinates from undistorted images to the orthomosaic. This process uses camera exterior orientation parameters and terrain elevation data from the “reconstruction.json” file and digital elevation model (DEM) generated by OpenSfM (\url{https://opensfm.org/}, accessed on February 14, 2025) during the ODM process in Module 2 (\autoref{fig:dia1}). Geospatial raster operations and affine transformations are performed using Rasterio (\url{https://rasterio.readthedocs.io/en/stable/}, accessed on February 14, 2025). This transformation corrects geometric distortions caused by camera tilt and terrain variation, ensuring spatially accurate plant positioning on the orthomosaic.

\begin{figure}[htbp]
  \centering
  \includegraphics[width=0.5\textwidth, height=0.5\textheight, keepaspectratio]{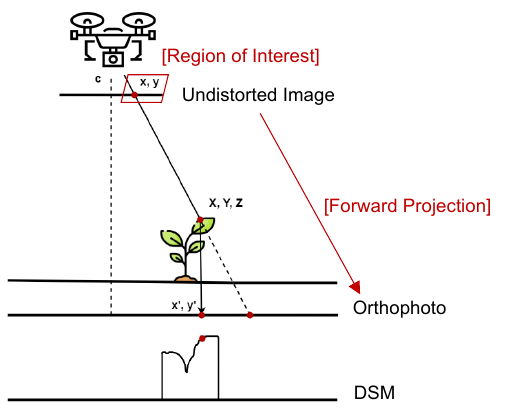}
  \caption{Forward projection of detected plant locations from undistorted images to the orthomosaic. The region of interest (ROI) localizes plant detections in the undistorted image, optimizing computational efficiency. The forward projection process transforms these detections into geospatial coordinates on the orthomosaic using camera parameters and digital surface model (DSM) data.}
  \label{fig:project}
\end{figure}

The algorithm in this module is developed to improve computational efficiency. A region of interest (ROI) is defined around each detection in the undistorted image, limiting the search space for the projection (\autoref{fig:project}). The corrected plant positions can then be reliably aligned with geospatial datasets for downstream phenotypic analysis.

\subsubsection{Module 9:Spatial Data Conversion}
Module 9 enables the conversion of projected bounding boxes into shapefiles and allows the extraction of spatial traits from raster data. Bounding box coordinates are transformed into rectangular polygons and stored with an appropriate coordinate reference system (CRS). The shapefiles are compatible with field-scale data such as canopy height models (CHMs) or vegetation index (VI) maps, such as the normalized difference vegetation index (NDVI) \autocite{sangjan2022optimization}. Since these datasets share a common CRS derived from the same GCPs, spatial overlays can be performed directly

\begin{figure}[htbp]
  \centering
  \fbox{\includegraphics[width=0.95\textwidth, height=0.95\textheight, keepaspectratio]{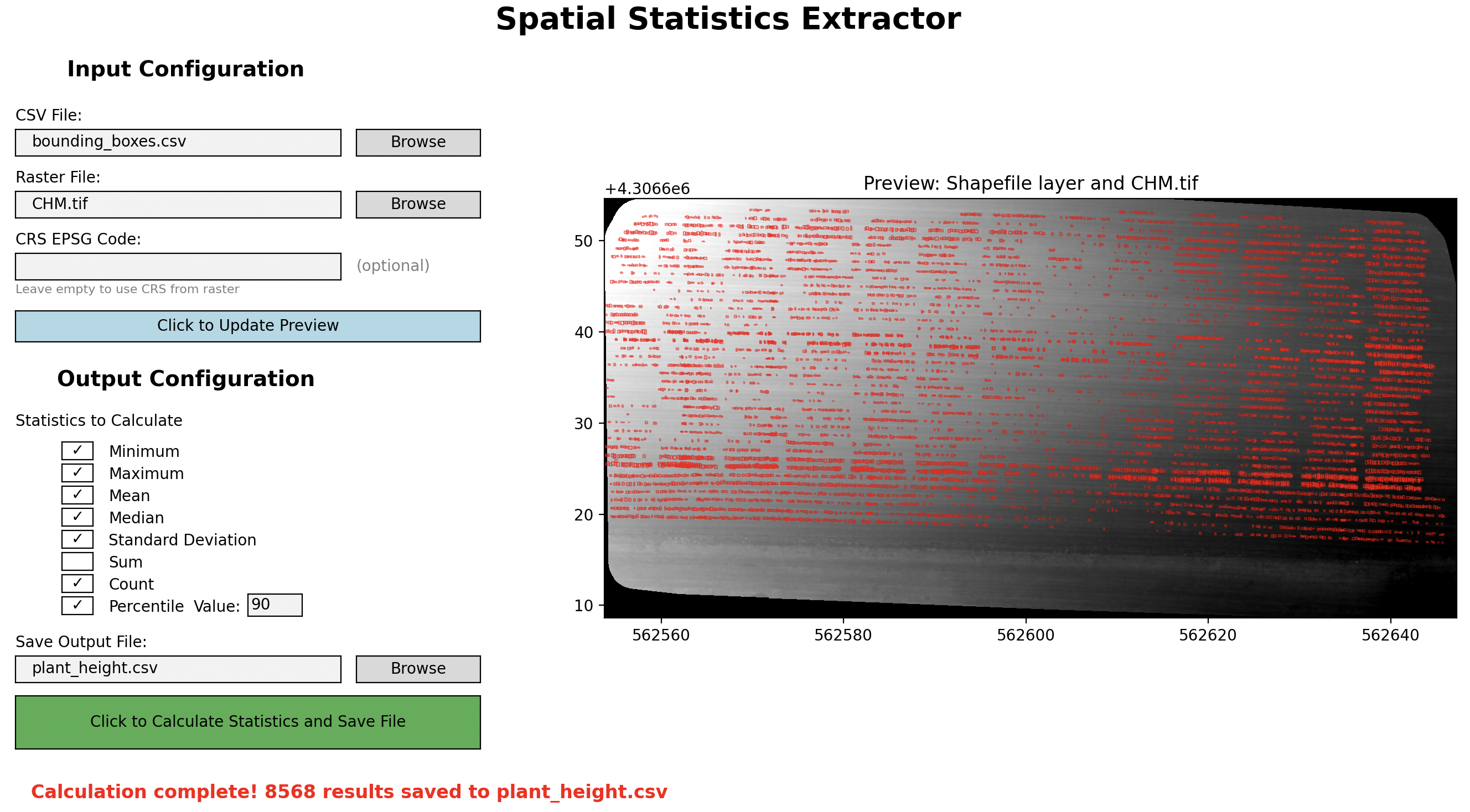}}
  \caption{GUI of the “9\_spatial\_stat\_extractor” module for extracting spatial statistics from georeferenced UAV data. Left side: file/raster selection and statistics configuration. Right side: shapefile preview over raster.}
  \label{fig:mod9}
\end{figure}

An interactive GUI was developed mainly using Matplotlib. It allows users to select inputs, including the bounding box CSV file, a georeferenced raster, and CRS configuration (\autoref{fig:mod9}). Users can choose the desired statistical values, while the preview panel shows the overlaid shapefile on the selected raster. Upon execution, the module computes zonal statistics for each polygon and exports the results to a CSV file. This functionality supports the efficient extraction of plant-level spectral and structural traits, enhancing the utility of UAV-based phenotyping workflows.

\section{Case Study Using MatchPlant for Individual Maize Detection}
\subsection{Data Collection}
RGB images were collected using a DJI Mavic 2 Pro with a 20-megapixel L1D-20c Hasselblad camera (SZ DJI Technology Co., Ltd., Shenzhen, China). Flights were conducted on a maize field experiment involving diverse maize inbred lines grown at the University of Missouri Genetics Farm in Columbia, MO, USA, on June 16 and June 23, 2021, corresponding to three and four weeks after planting, respectively. All Flights were executed at 20 m altitude with 85\% front and side overlap and during solar noon (11 AM to 2 PM local time) to maintain uniform illumination and minimize shadow effects. Each image had a resolution of 5472~$\times$~3648 pixels. Multispectral images were acquired using a DJI Matrice 600 (SZ DJI Technology Co., Ltd., Shenzhen, China) with a 1.2-megapixel RedEdge-MX camera (MicaSense Inc., Seattle, WA, USA), collected four weeks after planting using the same flight parameters. 

\subsection{Data Preprocessing}
\begin{figure}[htbp]
  \centering
  \includegraphics[width=\textwidth, height=\textheight, keepaspectratio]{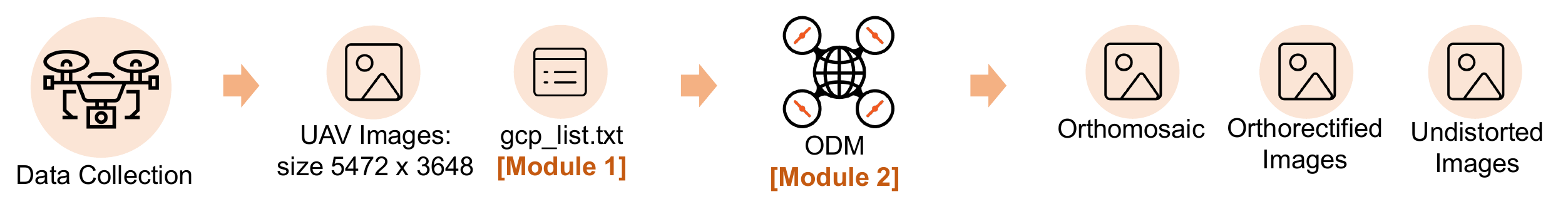}
  \caption{A diagram illustrating the data preprocessing for individual maize detection.}
  \label{fig:dia1}
\end{figure}

Raw UAV imagery was processed to generate georeferenced products for object detection and analysis. Module 1: GCP Finder identified GCPs and generated the “gcp\_list.txt” file. This was followed by Module 2: UAV Image Processing, where photogrammetric reconstruction was performed using the ODM pipeline via an automated script “run\_ODM\_process.sh” (or *.ps1) executed through Docker (\autoref{fig:dia1}). The key outputs included orthomosaic (ground sampling distance (GSD): 0.3 cm), orthorectified images, and undistorted images, which were used in downstream modules

\subsection{Data Preparation}
\begin{figure}[htbp]
  \centering
  \includegraphics[width=\textwidth, height=\textheight, keepaspectratio]{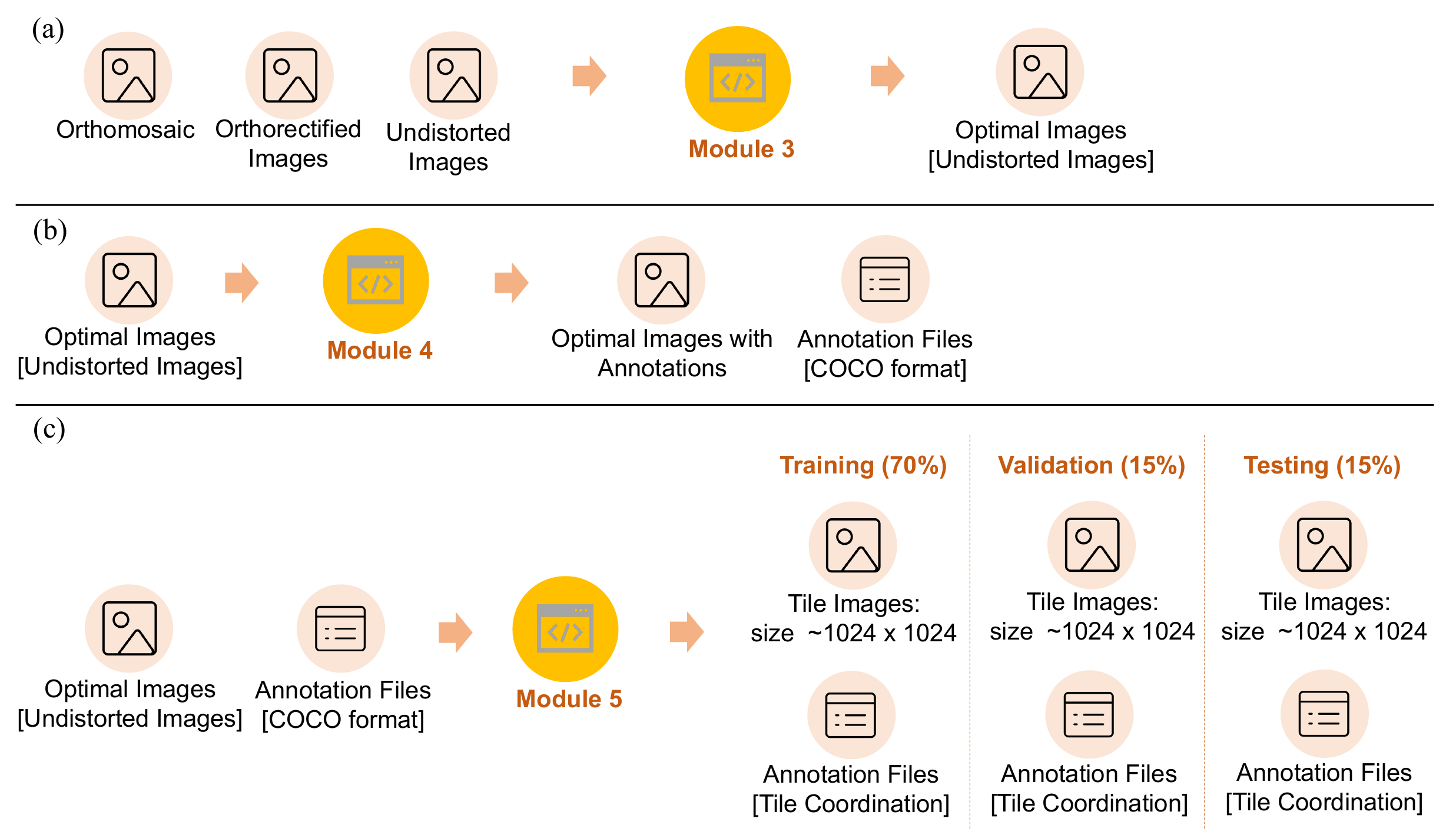}
  \caption{Diagrams illustrating the data preparation for individual maize detection. (a) Module 3-Minimum image selection process. (b) Module 4-Annotating undistorted images process. (c) Module 5-Tiling and splitting image dataset process.}
  \label{fig:dia2}
\end{figure}

In this case study, UAV imagery from the third-week post-planting was refined into a structured dataset for maize plant detection. Module 3: Image Selection reduced redundancy by selecting the minimum undistorted images needed for the full maize field coverage (\hyperref[fig:mod345]{Figure 3b} and \hyperref[fig:dia2]{Figure 7a}). Module 4: Bounding Box Annotation was followed, where maize plants were manually labeled using a custom-built GUI (\hyperref[fig:mod345]{Figures 3c and 3d}, and \hyperref[fig:dia2]{Figure 7b}). Annotations were created in the COCO (.json) with a single category used for maize.

Next, Module 5: Image Tiling and Dataset Splitting divided the annotated images into $\sim$1024~$\times$~1024 pixel tiles and recalculated bounding box coordinates. The dataset was then split into training (70\%), validation (15\%), and test (15\%) sets (\hyperref[fig:mod345]{Figures 3e and 3f}, and \hyperref[fig:dia2]{Figure 7c}). This structured data preparation improved annotation consistency and enabled seamless training of the object detection model. 

\subsection{Model Training and Validation}
\begin{figure}[htbp]
  \centering
  \includegraphics[width=\textwidth, height=\textheight, keepaspectratio]{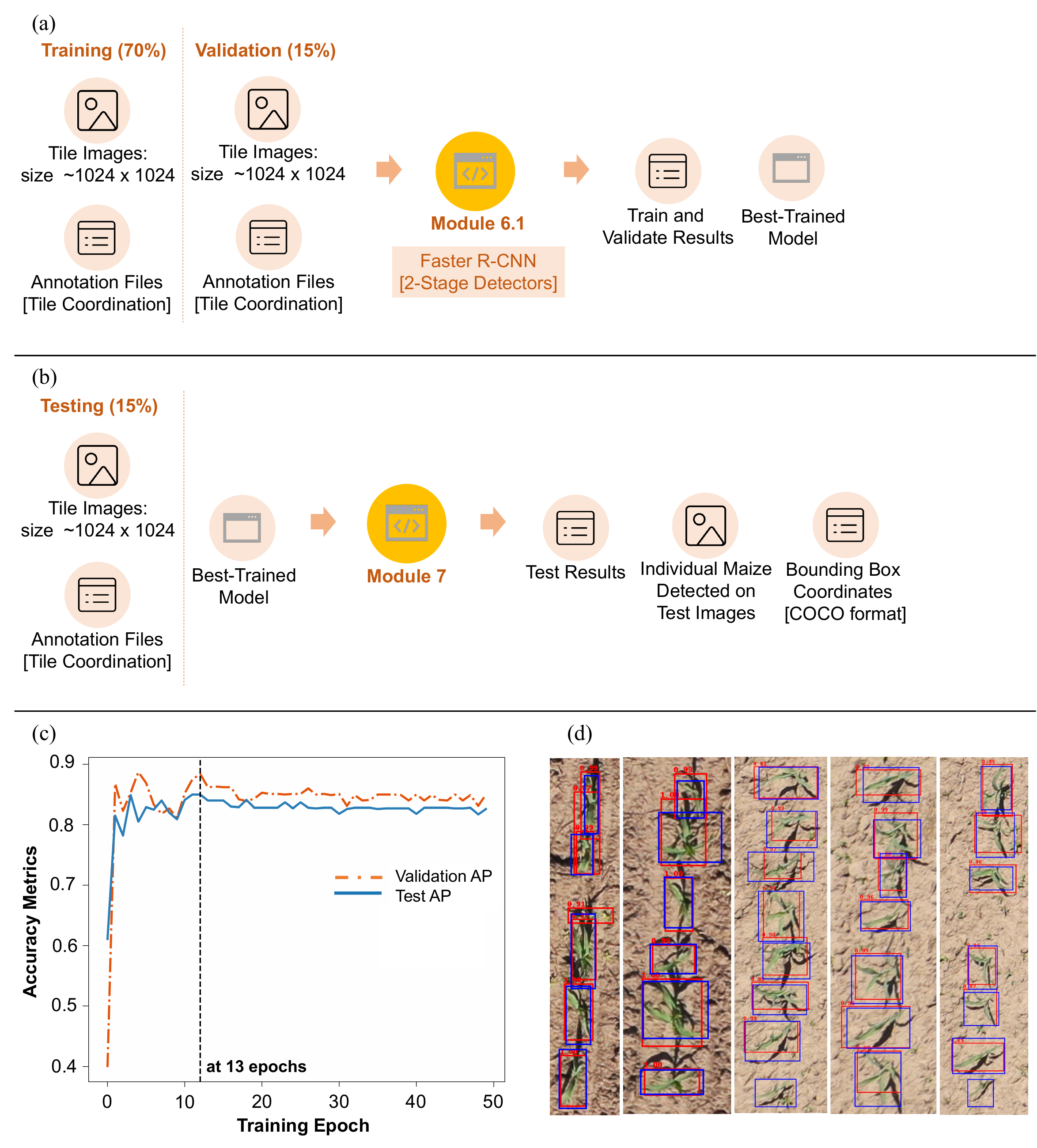}
  \caption{Model development for individual maize detection. (a) Module 6.1-Training process. (b) Module 7-Testing process. (c) Epoch-by-epoch evaluation of average precision (AP) at an intersection over the union (IoU) threshold = 0.5. Test AP was obtained using a separate evaluation script outside the training pipeline for comparison purposes. (d) Example test images with bounding boxes.}
  \label{fig:dia3}
\end{figure}

Module 6.1: Model Training used the prepared dataset to train a Faster R-CNN model (\hyperref[fig:dia3]{Figure 8a}). The training was configured via “train\_config.yaml” to specify a backbone learning rate of 0.005 and 0.01 for the RPN and ROI heads, respectively, using SGD with a momentum of 0.9 and a weight decay of 0.0005. A learning rate decay factor of 0.1 was applied every 10 epochs to optimize convergence. 

A confidence threshold of 0.05 was used during training, while Module 7: Model Testing applied a 0.5 threshold during inference to balance recall and precision. Epoch-wise validation and an external evaluation script on the test dataset revealed three key outcomes at IoU = 0.5 (\hyperref[fig:dia3]{Figure 8c}): (1) performance plateaus at epoch 13, indicating diminishing returns beyond this point; (2) a modest performance gap between validation average precision (AP) of 0.5 (89.6\%) and test AP of 0.5 (85.0\%) likely due to the test set contains more challenging examples; and (3) consistent test performance despite fluctuations in validation metrics, highlighting reliable generalization ability. 

Throughout the training, classification loss, bounding box regression loss, objectness loss, and RPN loss were tracked. Metrics and performance statistics were stored in the file “training\_summary.json.” The best-performing model at epoch 13 was saved with periodic checkpoints every five epochs.

\subsection{Model Testing and Evaluation}
The best-trained model was evaluated using five independent test runs in Module 7: Model Testing (\hyperref[fig:dia3]{Figure 8b}). The testing framework, configured via “test\_config.yaml,” the evaluation used a confidence threshold of 0.5. It returned an AP of 85.9\% at IoU = 0.5 and a mean average precision (mAP) of 40.4\% across IoU thresholds from 0.5 to 0.95.
Size-specific evaluation based on COCO object size categories (small:$~<$~1024 pixels$^{2}$, medium: 1024–9216 pixels$^{2}$, large:$~>~$9216 pixels$^{2}$) revealed the model performed best on large objects (97.0\% precision), followed by medium (87.7\%), and was least effective on small objects (31.9\%). 
All five test runs yielded consistent performance, confirming model robustness. Output included detection metrics, confidence distributions, and COCO-format prediction files (x, y, width, height) for downstream geospatial integration. 

\subsection{Utilization: Detected Plant Projection}
\begin{figure}[htbp]
  \centering
  \includegraphics[width=\textwidth, height=\textheight, keepaspectratio]{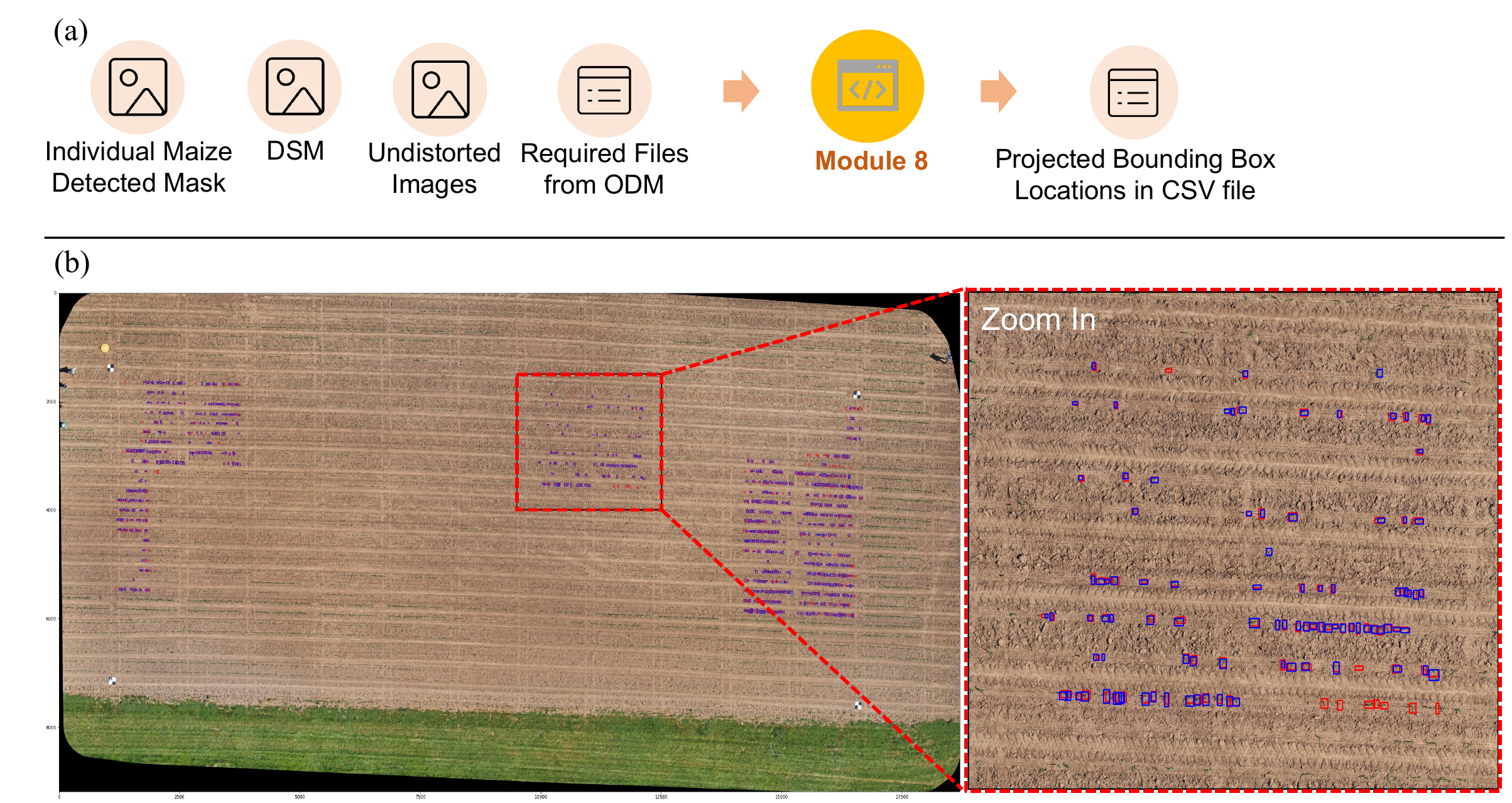}
  \caption{Projection of detected maize plant locations onto the orthomosaic. (a) Module 8-Detected plant projection. (b) Left: orthomosaic with projected bounding boxes. Right: zoomed view comparing projected detections(blue) and manual annotation (red).}
  \label{fig:dia4}
\end{figure}

The prediction file contained predicted bounding box locations from Module 7, which were transformed into binary masks matching the dimensions of the undistorted source image. In Module 8: Detected Plant Projection, these masks, undistorted images, digital surface model (DSM), and metadata from ODM were used to project detections onto an orthomosaic using forward projection (\autoref{fig:project} and \hyperref[fig:dia4]{Figure 9a}).

For evaluation, the bounding boxes identified in the test set images were projected onto the orthomosaic (\hyperref[fig:dia4]{Figure 9b}). Of the 964 manually drawn test-set bounding boxes, 866 were projected successfully, yielding an 89.8\% georeferencing rate. IoU analysis showed that 87.5\% of projections matched manual annotations at IoU = 0.5, confirming spatial alignment. Final results were exported in CSV (comma-separated values) format for GIS-based plant-level analysis.

\subsection{Utilization: Spatial Data Conversion}
\begin{figure}[htbp]
  \centering
  \includegraphics[width=\textwidth, height=\textheight, keepaspectratio]{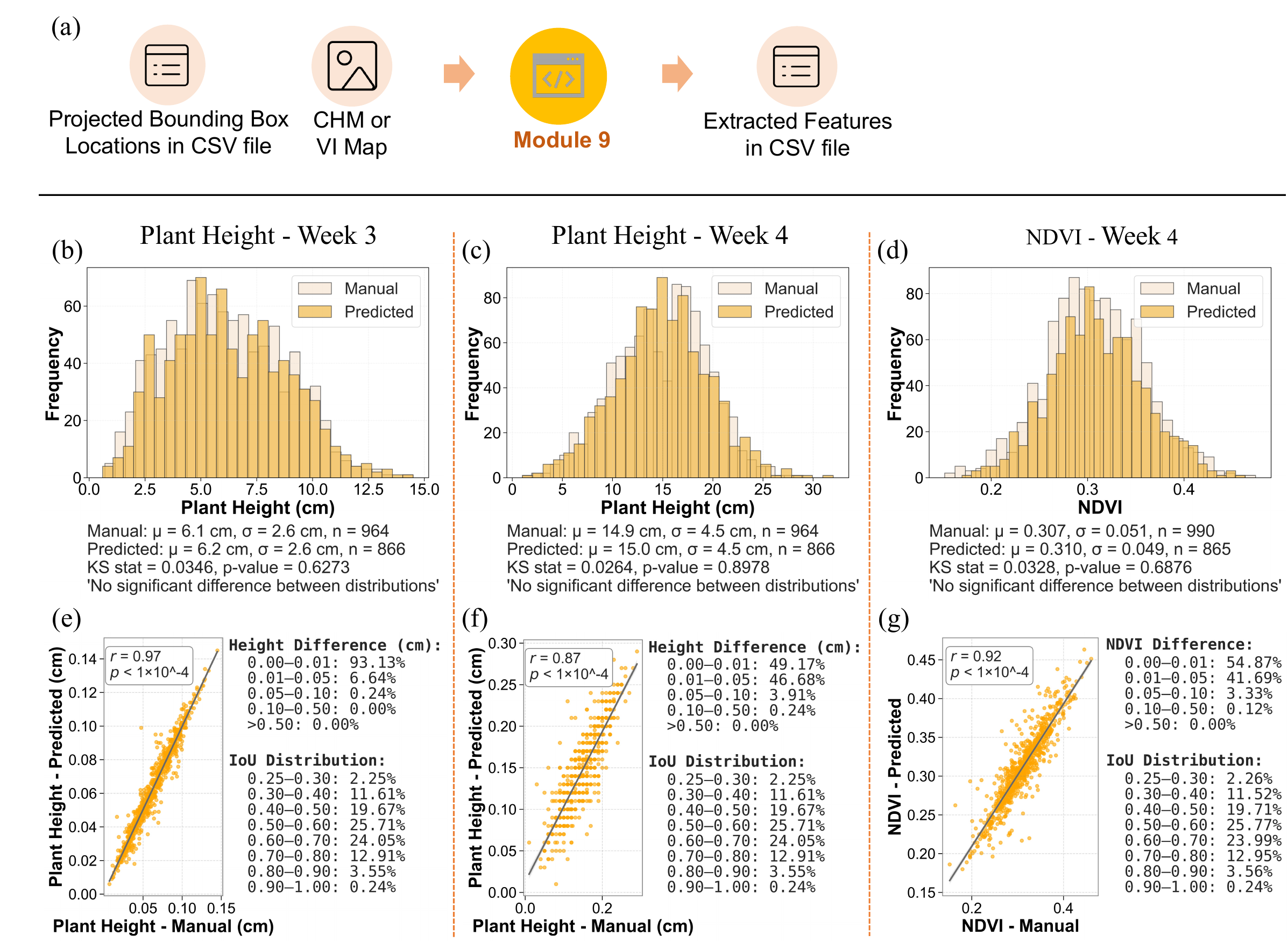}
  \caption{Phenotypic trait extraction and validation. (a) Module 9-Spatial Data Conversion. (b–d) Distribution comparisons of plant height [$\mu$: mean, $\sigma$: standard deviation, KS: Kolmogorov-Smirnov–Smirnov]. (e–g) Correlation plots with summaries of absolute trait differences and IoU evaluation. }
  \label{fig:dia5}
\end{figure}

Module 9: Spatial Data Conversion converted projected bounding boxes (CSV file) into shapefiles (\hyperref[fig:dia5]{Figure 10a}). These shapefiles were aligned with UAV-derived CHM in Weeks 3 and 4 after planting, following a method adapted from \textcite{tirado2020uav}. Additionally, NDVI maps derived from multispectral UAV imagery in Week 4 were incorporated into the analysis. 

Plant-level phenotypic traits (plant height and NDVI in this case study) were extracted from predicted bounding box shapefiles. The extracted values were compared to those from manual annotations to assess the reliability of automated object detection.

Statistical analyses using SciPy (\url{https://scipy.org/}, accessed on February 14, 2025) confirmed the high agreement between the two approaches. Distribution comparisons using the Kolmogorov–Smirnov test \autocite{hodges1958significance} showed no significant distribution differences ($p$ $>$ 0.05) (\hyperref[fig:dia5]{Figure 10b-d}). Correlation analysis revealed high linear relationships between the two extraction methods ($r$ = 0.87 to 0.97) (\hyperref[fig:dia5]{Figure 10e-g}). 

Computed IoU values mostly exceeded 0.4–0.5, indicating consistent spatial alignment. These results demonstrate the effectiveness of the MatchPlant pipeline for automated trait extraction in UAV-based phenotyping workflows.

\section{Discussion}
MatchPlant aims to broaden the applications in UAV-based imagery for individual plant detection and georeferenced trait extraction through a modular, open-source pipeline. This section highlights key considerations, limitations, and future directions to support its integration into agricultural practice and research workflows.

\subsection{Data Preprocessing}
Modules 0 and 1, responsible for GPS embedding and GCP file creation, contribute to spatial accuracy, which is an essential requirement for precise plant-level phenotyping. Integrating ODM as an open-source photogrammetric engine (Module 2) enhances pipeline accessibility by removing licensing constraints associated with commercial software. The generation of undistorted imagery provides higher-quality inputs for object detection training. 

Despite these advantages, ODM presents limitations relative to commercial alternatives, including slower processing times under default settings, limited support for multispectral calibration, and reduced efficiency when handling large datasets \autocite{pell2022demystifying}. However, commercial tools often lack access to intermediate orthorectification data, such as undistorted and orthorectified images, which are required for matching raw image coordinates with orthomosaic coordinates. In contrast, ODM provides access to its processing pipeline and intermediate outputs, making it particularly well-suited for MatchPlant workflows that rely on georeferenced object projection. 

Orthorectification accuracy in ODM, as with any photogrammetric solution, is also influenced by the quantity and spatial distribution of GCPs. Especially in complex terrain, suboptimal GCP placement can lead to spatial misalignments that propagate through the projection, trait extraction, and phenotyping analysis steps \autocite{pugh2021comparison, mora2024performance}.

\subsection{Data Preparation}
The Image Selection Module 3 reduces dataset redundancy in high-overlap UAV missions by filtering images based on spatial coverage. However, its reliance on orthorectified metadata may lead to inconsistencies when GCPs are sparse and misaligned. The GUI-based annotation Module 4 provides an intuitive GUI that supports COCO and YOLO output formats, chosen for their broad compatibility with DL frameworks. For users requiring advanced features, external platforms such as Roboflow or CVAT (\url{https://roboflow.com/}; \url{https://www.cvat.ai/}, accessed on February 14, 2025) remain compatible. Nonetheless, the built-in tool offers a lightweight, offline solution that preserves data privacy and simplifies the workflow. 

The image tiling Module 5 is designed to reduce the risk of splitting objects at tile boundaries by minimizing object fragmentation through configurable overlap and IoU thresholds. Although the GUI provides interactive overlap adjustments, careful tuning is still required. Inadequate overlap can reduce annotation completeness and impair model performance, particularly near tile boundaries.

\subsection{Object Detection Model-Training and Testing}
MatchPlant supports training and testing using Faster R-CNN with a ResNet-50 FPN backbone, selected for its ability to detect small objects in UAV imagery. While this choice simplifies deployment and supports robust performance, it may limit flexibility for users seeking alternative architectures. The model applies a dual-threshold strategy: a lower confidence threshold (0.05) during training improves recall, while a higher threshold (0.5) during testing increases precision, supporting robust object detection performance.

The transfer learning module provides additional flexibility by supporting selective layer freezing and component-specific learning rates. These features are beneficial in phenotyping tasks with limited annotated data. In the maize case study, the final model achieved an AP of 85.9\% at an IoU of 0.5; however, the mAP across IoU thresholds from 0.5 to 0.95 dropped to 40.4\%. This reduction occurs in UAV-based phenotyping, where small spatial misalignments, especially for early-stage crops, can lead to significant penalties under stricter IoU criteria \autocite{tang2024survey, wu2024dense, zhao2024object}.

Evaluation across object sizes revealed strong performance for medium and large objects but lower accuracy for very small instances ($<$~32~$\times$~32 pixels). Although the dataset focused on early-stage plants to reduce occlusion, its limited pixel footprint in UAV images affected detection reliability. Future improvements may include increasing image resolution via lower flight altitude, using higher-resolution sensors, or image enhancement techniques before training \autocite{varol2025novel, zhou2025multi}.

\subsection{Utilization of Detection Outputs}
Modules 8 and 9 connect object detection outputs with geospatial analysis by projecting detections onto orthomosaic and converting them into GIS-compatible shapefiles. These processes facilitate the automated extraction of plant-level traits from CHM and VI maps, supporting phenotyping workflows.

Comparison of the statistical distribution of manually derived and automatically extracted traits (\hyperref[fig:dia5]{Figures 10b–d}) showed a high relation in both mean and variance for plant height (Weeks 3 and 4) and NDVI (Week 4), confirming the reliability of the projected bounding boxes for downstream phenotypic analysis. Correlation plots and IoU distributions (\hyperref[fig:dia5]{Figures 10e–g)} further validated spatial precision, with linear relationships ($r$ = 0.87–0.97) and acceptable spatial overlap (IoU $>$ 0.4) between prediction and manual annotations. 

A key advantage is the temporal reusability of georeferenced bounding boxes. In the case study, projected bounding boxes from Week 3 were reused to extract plant traits in Week 4, minimizing annotation effort. However, plant growth introduced minor spatial shifts, indicating the need to account for structural variability over time. One solution is to dynamically integrate CHM-derived height data to refine bounding box placement. Though not yet implemented, this approach could enhance temporal consistency without requiring new annotation. Enhancing the GUI and integrating uncertainty metrics may improve the accessibility and interpretability of projection outputs in the phenotyping pipeline.

\subsection{Algorithm Development}
MatchPlant is primarily implemented in Python, utilizing widely adopted libraries such as NumPy, OpenCV, Rasterio, and PyTorch. GUI modules are built using Matplotlib, providing a lightweight, platform-independent solution without external frameworks. However, interactivity and responsiveness remain constrained by Matplotlib’s native limitations, which may restrict the development of more advanced GUI features available in toolkits such as PyQt or Tkinter (\url{https://www.riverbankcomputing.com/software/pyqt/}; \url{https://docs.python.org/3/library/tk.html}, accessed on February 14, 2025). 

The codebase is distributed as modular Python scripts, enabling customization and reuse of components for various object detection tasks. However, this script-based deployment requires users to configure a Python environment with the necessary dependencies. While this approach supports transparency and reproducibility, it may pose a barrier for non-technical users. Future development will prioritize packaged executables for Windows and macOS to improve accessibility.

Manual parameter tuning and the absence of integrated error handling or logging may hinder pipeline robustness at scale. Addressing these gaps and supporting additional model architectures will be key goals for improving MatchPlant’s usability and adaptability across research domains.

\section{Conclusion}
This study introduces MatchPlant, a modular, open-source pipeline designed to support UAV-based individual plant detection and spatial trait extraction. By integrating image preprocessing, data preparation, deep learning-based detection, geospatial projection, and trait extraction, MatchPlant enables end-to-end phenotyping workflows suitable for plant phenotyping applications.

The pipeline’s use of undistorted imagery for model training enhances object detection accuracy, while its georeferenced projection module ensures spatial consistency with downstream datasets. In the presented maize case study, MatchPlant demonstrated high detection accuracy and strong agreement between projected bounding boxes and manual annotations. Additionally, the ability to reuse earlier-season detections across later time points reduced annotation demands while maintaining spatial alignment, with only minor discrepancies attributed to plant growth. However, these time points were all early in the season, and extending them to mid- or late-season may require additional method development.

MatchPlant offers a flexible and extensible foundation for integrating UAV imagery into high-throughput phenotyping pipelines. Future work will explore bounding box refinement using CHM data, support for additional detection models, and packaging enhancements to improve accessibility for a broader user base.

\section*{CRediT authorship contribution statement}
\textbf{Worasit Sangjan}: Conceptualization, Methodology, Software, Validation, Formal analysis, Investigation, Data curation, Writing – original draft, Writing – review \& editing, Visualization. \textbf{Piyush Pandey}: Conceptualization, Methodology, Formal analysis, Writing – review \& editing. \textbf{Norman B. Best}: Conceptualization, Resources, Writing – review \& editing, Supervision, Project administration, Funding acquisition. \textbf{Jacob D. Washburn}: Conceptualization, Resources, Writing – review \& editing, Supervision, Project administration, Funding acquisition.

\section*{Declaration of Competing Interest}
The authors declare that they have no known competing financial interests or personal relationships that could have appeared to influence the work reported in this paper.

\section*{Data availability}
The public datasets supporting the case study are available on Zenodo at \url{https://doi.org/10.5281/zenodo.14856123} (accessed on February 14, 2025). The source code and documentation for MatchPlant are available on GitHub at \url{https://github.com/JacobWashburn-USDA/MatchPlant} (accessed on February 14, 2025).  

\section*{Acknowledgments}
This research was supported in part by an appointment to the Agricultural Research Service (ARS) Research Participation Program administered by the Oak Ridge Institute for Science and Education (ORISE) through an interagency agreement between the U.S. Department of Energy (DOE) and the U.S. Department of Agriculture (USDA). ORISE is managed by ORAU under DOE contract number DE-SC0014664. Funding was also provided by the USDA Agricultural Research Service, SCINet Postdoctoral Fellows Program. This research used resources provided by the SCINet project and/or the AI Center of Excellence of the USDA Agricultural Research Service, ARS project numbers 0201-88888-003-000D and 0201-88888-002-000D.  All opinions expressed in this publication are the author’s and do not necessarily reflect the policies and views of USDA, DOE, or ORAU/ORISE.

\begin{sloppy}
\printbibliography
\end{sloppy}

\end{spacing} 
\end{document}